\documentclass[sigconf]{acmart}
\usepackage{booktabs}
\usepackage[title]{appendix}
\usepackage{makecell}
\usepackage{csquotes}
\usepackage{xcolor}
\usepackage{subcaption}
\usepackage{graphicx}
\usepackage{tcolorbox}
\usepackage{multirow}
\usepackage{url}
\usepackage{hyperref}
\renewcommand\footnotetextcopyrightpermission[1]{} 

\begin{document}
%%%%%%%%%%%%%%%%%%%%%%%%%%%%%%%%

\pagestyle{empty}

%%%%%%%%%%%%%%%%%%%%%%%%%%%%%%%%
\title{Stress Testing BERT Anaphora Resolution Models for Reaction Extraction in Chemical Patents}
%%%%%%%%%%%%%%%%%%%%%%%%%%%%%%%%

\author{\textbf{Chieling Yueh}}
\authornote{Corresponding author}
\affiliation{
  \institution{University of Amsterdam \& Elsevier}
  \city{Amsterdam} 
  \country{The Netherlands}
  }
\email{chielingyueh@gmail.com}

\author{\textbf{Evangelos Kanoulas}}
\affiliation{
  \institution{University of Amsterdam}
  \city{Amsterdam} 
  \country{The Netherlands}
  }
\email{e.kanoulas@uva.nl}

\author{\textbf{Bruno Martins}}
\affiliation{
  \institution{Elsevier}
  \city{Amsterdam} 
  \country{The Netherlands}
  }
\email{b.martins@elsevier.com}

\author{\textbf{Camilo Thorne}}
\affiliation{
  \institution{Elsevier}
  \city{Frankfurt} 
  \country{Germany}
  }
\email{c.thorne.1@elsevier.com}

\author{\textbf{Saber Akhondi}}
\affiliation{
  \institution{Elsevier}
  \city{Amsterdam} 
  \country{The Netherlands}
  }
\email{s.akhondi@elsevier.com}
%%%%%%%%%%%%%%%%%%%%%%%%%%%%%%%%

%%%%%%%%%%%%%%%%%%%%%%%%%%%%%%%%
\begin{abstract}
The high volume of published chemical patents and the importance of a timely acquisition of their information gives rise to automating information extraction from chemical patents.
Anaphora resolution is an important component of comprehensive information extraction, and is critical for extracting reactions.
In chemical patents, there are five anaphoric relations of interest: co-reference, transformed, reaction associated, work up, and contained \cite{Dutt2021APatents, Fang2021ChEMU-Ref:Domain, Li2021ExtendedPatents}.
Our goal is to investigate how the performance of anaphora resolution models for reaction texts in chemical patents differs in a noise-free and noisy environment and to what extent we can improve the robustness against noise of the model.%
\footnote{%
GitHub:~~
\href{https://github.com/chielingyueh/anaphora_resolution_chemical_patents}{\url{https://github.com/chielingyueh/anaphora_resolution_chemical_patents}}
}
\end{abstract}
%%%%%%%%%%%%%%%%%%%%%%%%%%%%%%%%

%%%%%%%%%%%%%%%%%%%%%%%%%%%%%%%%
%%%%%%%%%%%%%%%%%%%%%%%%%%%%%%%%

\maketitle
%%%%%%%%%%%%%%%%%%%%%%%%%%%%%%%%
\section{Introduction}
\label{intro}
%%%%%%%%%%%%%%%%%%%%%%%%%%%%%%%%

Chemical patents are usually the first public disclosure of a new chemical compound \cite{Zhai2019ImprovingEmbeddings}. However, it can take one to three years for a chemical compound to be mentioned in scientific literature \cite{ Senger2015ManagingPatents}.
Moreover, chemical patents are crucial as a starting point in the understanding of a chemical compound's potential application and its properties \cite{Dutt2021APatents, Zhai2019ImprovingEmbeddings}, making them an important source of information \cite{Senger2015ManagingPatents}.
A large number of chemical patents are available, which makes manual information extraction of these patents time-consuming and costly \cite{Li2021ExtendedPatents, Akhondi2019AutomaticPatents}. As a result, current research has focused on automating information extraction from chemical patents \cite{Li2021ExtendedPatents, He2021ChEMUPatents, Krallinger2015CHEMDNER:Challenge, Dutt2021APatents, Fang2021ChEMU-Ref:Domain}.

A key component of comprehensive information extraction is \textit{anaphora resolution }\cite{Rosiger2019ComputationalResolution, Fang2021ChEMU-Ref:Domain}. Anaphora resolution is a linguistic phenomenon, in which an entity, i.e., the \emph{anaphor}, refers back to a previously mentioned entity, i.e., its \emph{antecedent} \cite{Mitkov2003TheLinguistics}. Anaphora resolution can be divided into two tasks, namely \textit{co-reference} and \textit{bridging resolution} \cite{Rosiger2019ComputationalResolution}. Co-reference resolution is a task, in which we want to identify all expressions in a text that refer to the same entity \cite{Clark2016ImprovingRepresentations}. Hence, in co-reference resolution, the anaphor refers to an antecedent which is identical \cite{Rosiger2019ComputationalResolution}. In bridging resolution, the relation between the anaphor and its referring antecedent is a non-identical association \cite{Hou2018UnrestrictedResolution, Kobayashi2020BridgingArt}.

\begin{figure*}[h]
  \centering
  \includegraphics[width=\linewidth]{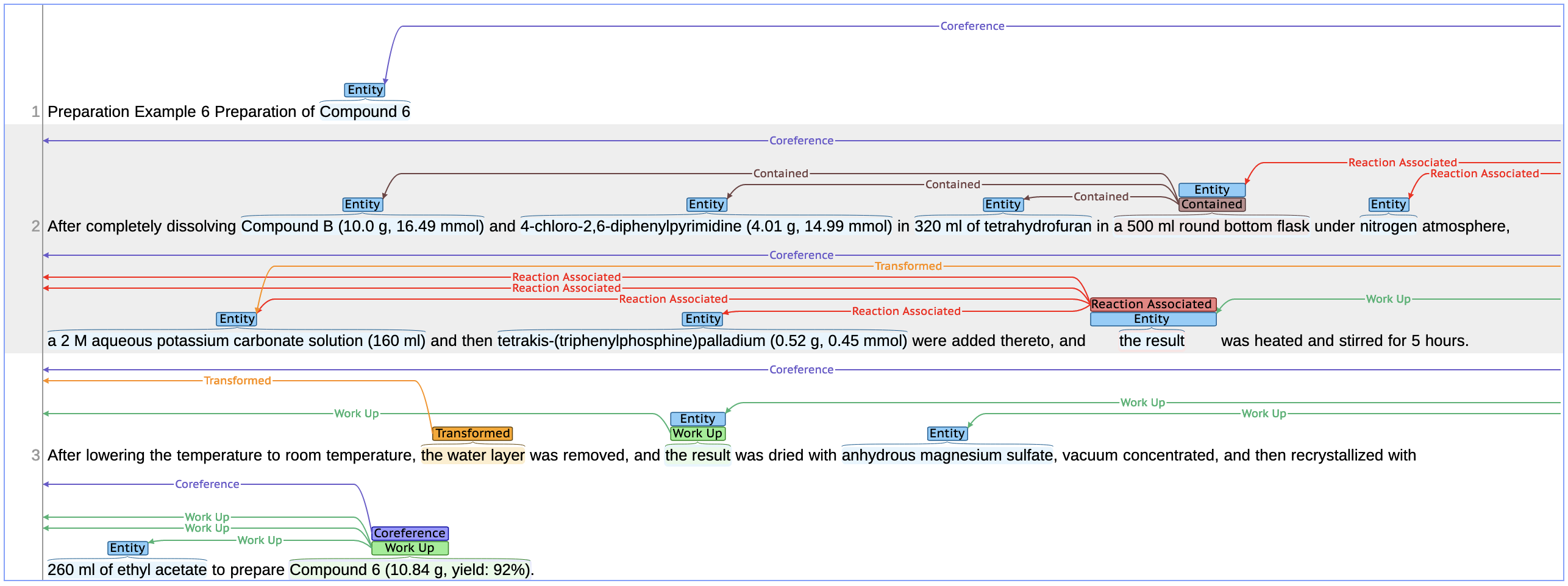}
  \caption{Annotated snippet of a chemical patent. Different colours of links represent
different anaphoric relation types.}
  \label{fig:AR_example}
\end{figure*}

Anaphora resolution is a critical -- though until recently, not well understood -- step for extracting reactions from chemical documents, and in particular chemical patents. Recent prior work on anaphora resolution in chemical patents \cite{Dutt2021APatents, Fang2021ChEMU-Ref:Domain, Li2021ExtendedPatents} enabled the definition of five different types of anaphoric relations -- \textit{co-reference}, \textit{transformed}, \textit{reaction associated}, \textit{work up} and \textit{contained} (see \autoref{tab:def_AR}), in which the latter four are bridging relations.
In the context of chemical patents, a co-reference relation is a relation in which different surface mentions refer to the same chemical entity \cite{Dutt2021APatents}. For example, "\textit{Compound 6}" in line 1 and "\textit{Compound 6 (10.85 g, yield 92\%)}" in line 3 of \autoref{fig:AR_example} refer to the same chemical entity. Therefore, these entities have a \textit{co-reference} relation. Bridging relations in chemical patents are relations in which different entities interact amongst themselves in a particular manner \cite{Dutt2021APatents}. For example, "\textit{Compound B (10.0 g, 16.49 mmol)}" and "\textit{a 500 ml round bottom flask}" in line 2 of  \autoref{fig:AR_example} have an interaction between each other, in which the first entity is \textit{contained} in the second entity.  \autoref{fig:AR_example} shows an annotated snippet of a chemical patent with examples of the different anaphoric relations in chemical patents. 

\begin{table}
  \caption{Definitions of the anaphoric relations in chemical patents \cite{Fang2021ChEMU-Ref:Domain}.}
  \label{tab:def_AR}
  \begin{tabular}{ll}
    \toprule
    {\textbf{Relation}} & {\textbf{Definition}} \\
    \midrule
    Co-reference & \makecell[l]{Relation between mentions/entities \\
    in which they share the same \\ chemical  
    properties.}\\
    \midrule
    Transformed & \makecell[l]{Relation between mentions/entities \\
    which are based on the same \\ 
    chemical components but have \\
    undergone a change in condition.}\\
    \midrule
    Reaction associated & \makecell[l]{Relation between a chemical \\
    compound and its immediate \\ source compounds via a mixing \\
    process in which the source \\
    compounds maintain their original \\
    chemical structure.}\\
    \midrule
    Work up & \makecell[l]{Relation between chemical \\ 
    compounds in which they were \\
    used to isolate or purify an \\
    associated output product. } \\
    \midrule
    Contained & \makecell[l]{Relation between a chemical \\ 
    compound and the equipment in \\
    which they are placed.}\\
    \bottomrule
  \end{tabular}
\end{table}

In this paper, we build further on the work of \cite{Dutt2021APatents}. Common errors made by the named entity recognition (NER) model in the mention extraction phase were span boundary mistakes \cite{Dutt2021APatents}. An example of a span boundary mistake is when the NER predicted "\textit{[mixture and]}" as an entity while the true entity is "\textit{[mixture]}".
Span boundary mistakes were minimised by doing a post-processing step after the mention extraction step \cite{Dutt2021APatents}. While they showed that the post-processing improved the performance of the mention extraction step, they did not show how the span boundary mistakes affect the relation classification step. In the rest of the paper, we refer to these span boundary mistakes as \textit{NER errors}.
Moreover, errors in the dataset can occur due to optical character recognition (\textit{OCR}) \textit{failures} \cite{Sayle2011ImprovedCorrection}. Chemical patents are usually in the format XML, HTML or image PDFs \cite{Akhondi2014AnnotatedMining}. For the image PDFs, OCR is used to extract the textual data from the PDFs. As a consequence, chemical patents can have typographical errors \cite{Sayle2011ImprovedCorrection}.
If no proper corrections are made, \textit{NER errors} \cite{Dutt2021APatents} and \textit{OCR failures} \cite{Sayle2011ImprovedCorrection} are likely to occur, resulting in noisy data for the relation classifier.

Thus, we aim to evaluate how the performance of the relation classification model on the task of anaphora resolution in chemical patents differs in a noise-free and noisy environment. Moreover, we aim to improve the robustness of the model against noise. To investigate this, we first develop a relation classification model to identify the anaphoric relations in chemical patents by replicating the relation classification step of \cite{Dutt2021APatents}. We evaluate our model in a noise-free environment. 
Secondly, we perform a stress test evaluation to evaluate our model in a noisy environment. 
Thirdly, we improve the robustness of our model by adding simulated OCR failures and NER errors to the training.
For simplicity, we focus on relation classification models only, rather than on end-to-end models (including NER).

%%%%%%%%%%%%%%%%%%%%%%%%%%%%%%%%
\section{Related Work}
%%%%%%%%%%%%%%%%%%%%%%%%%%%%%%%%

%%%%%%%%%%%%%%%%%%%%%%%%%%%%%%%%
\subsection{Relation Classification}
\label{subsec:rc}
%%%%%%%%%%%%%%%%%%%%%%%%%%%%%%%%

Relation classification is a natural language processing (NLP) task in which the semantic relationship between two mentions is predicted \cite{Wu2019EnrichingClassification}. 
Thus, the objective is to identify the relationship between two mentions, $Ent_1$ and $Ent_2$, in a text \cite{Hendrickx2010SemEval-2010Nominals}. For example, in the case of chemical patents, "\textit{[Compound A]}$_1$ was obtained as a \textit{[yellow solid]}$_2$", in which the entity pair "\textit{[Compound A]}$_1$" and "\textit{[yellow solid]}$_2$" should be classified as a \textit{co-reference} relation, as their chemical properties are the same.

The task of relation classification has been done using deep neural networks \cite{Socher2012SemanticSpaces, Zeng2014RelationNetwork, Shen2016Attention-BasedExtraction}. In particular, transformers-based models, such as BERT \cite{Devlin2019BERT:Understanding}, have achieved state-of-the-art results in many NLP tasks, such as question answering \cite{Zhang2020RetrospectiveComprehension}, sentiment analysis \cite{Bataa2019AnJapanese}, named entity recognition \cite{Hakala2019BiomedicalBERT} and summarisation \cite{Zhong2019SearchingNext}.
Likewise, transformer models have been tested on the task of relation classification, which achieved state-of-the-art results \cite{Wu2019EnrichingClassification, Soares2019MatchingLearning, Peng2019TransferDatasets, Eberts2019Span-basedPre-training, Dutt2021APatents}.
Thus, we use transformer models for our relation classification task.

Transformer models for relation classification have been improved by adding special tokens to the input. For example, the authors of \cite{Wu2019EnrichingClassification} used BERT for the relation classification task, in which they used special tokens for their input sentences. They appended "[CLS]" at the beginning of a sentence, "[SEP]" to separate two sentences, "\$" for the first entity and "\#" for the second entity. The "\$" and "\#" tokens help the model identify the entities' locations and transfer that knowledge into BERT, in which BERT's model output contains the location information of both entities. The "[SEP]" token is used when a relation between two entities spans across multiple sentences. The work of \cite{Wu2019EnrichingClassification} compared BERT models using special tokens and not using special tokens. The performance in terms of F1 was higher for the model that used special tokens. 

Yet, another approach is based on using the locations of entities in BERT models in the task of relation classification \cite{Soares2019MatchingLearning}. 
The authors of \cite{Soares2019MatchingLearning} used entity markers tokens, [$E1_{start}$], [$E1_{end}$], [$E2_{start}$] and [$E2_{end}$], to indicate the beginning and ending of the entities. 
As in \cite{Wu2019EnrichingClassification}, the authors of \cite{Soares2019MatchingLearning} found that adding the entity markers to the input of BERT outperformed the BERT model that did not use the entity markers. 

As \cite{Wu2019EnrichingClassification} and \cite{Soares2019MatchingLearning} both showed that introducing these entity markers to the model improved the performance in comparison to not using entity markers, our relation classification model uses entity markers to indicate the beginning and ending of the entities.

%%%%%%%%%%%%%%%%%%%%%%%%%%%%%%%%
\subsection{Anaphora Resolution in Chemical Patents}
%%%%%%%%%%%%%%%%%%%%%%%%%%%%%%%%

As mentioned in \autoref{intro}, prior research has been conducted on anaphora resolution in chemical patents \cite{Dutt2021APatents, Fang2021ChEMU-Ref:Domain, Li2021ExtendedPatents}. In \cite{Fang2021ChEMU-Ref:Domain}, they developed the ChEMU-ref dataset -- a corpus for modeling anaphora resolution in chemical patents. The ChEMU-ref dataset is created for the anaphora resolution task in the shared task organised by Cheminformatics Elsevier Melbourne Universities lab (ChEMU) \cite{Li2021ExtendedPatents}. ChEMU proposed a neural approach classifying the different anaphoric relations in the ChEMU-ref dataset in which they jointly trained co-reference and bridging relations, achieving an F1-score of 0.821 \cite{Li2021ExtendedPatents}.

As previously mentioned in \autoref{intro}, a pipelined approach to anaphora resolution in chemical patents is proposed by \cite{Dutt2021APatents} using the ChEMU-ref dataset. Their relation classification model outperformed the baseline of \cite{Li2021ExtendedPatents}, achieving a F1 score of 0.927, by ensembling the following BERT models: BERT-base \cite{Devlin2019BERT:Understanding}, BERT-large \cite{Devlin2019BERT:Understanding}, BioBERT \cite{Lee2020BioBERT:Mining}, Clinical-BERT \cite{Alsentzer2019PubliclyEmbeddings} and PubMed-BERT \cite{Peng2019TransferDatasets}.
A common error that was found were span boundary mistakes, resulted from the mention extraction phase.
An example of a span boundary mistake is when the model predicts "[Alcohol and]" as an entity, but the true entity is "[Alcohol]". 
We build further on the work of \cite{Dutt2021APatents} by evaluating how span boundary mistakes, referred as NER errors, impact the relation classification.

%%%%%%%%%%%%%%%%%%%%%%%%%%%%%%%%
\subsection{Stress Test Evaluation}
%%%%%%%%%%%%%%%%%%%%%%%%%%%%%%%%

NLP models can be tested by performing a stress test evaluation. Stress test evaluating is a way to test a model whether noisy data affect the performance \cite{Araujo2021StressEmbeddings}. Especially in neural computer vision models, it is shown that noisy data can have a severe impact on the performance \cite{Akhtar2018ThreatSurvey}.
The authors of \cite{Aspillaga2020StressTasks} did a stress test evaluation in NLP tasks, such as natural language inference and question answering. Their work found that the transformer models, RoBERTa, XLNet and BERT, are more robust against noisy data than recurrent neural networks for both tasks. 

As previously mentioned, due to OCR failures and NER errors, the input of the data can be noisy. For this reason, it is relevant to evaluate how the performance of our model is affected by such noisy data. Therefore, we perform a stress test evaluation on our model. 

%%%%%%%%%%%%%%%%%%%%%%%%%%%%%%%%
\section{Methodology}
\label{method}
%%%%%%%%%%%%%%%%%%%%%%%%%%%%%%%%

We develop a relation classification model using BERT to classify anaphoric relationships based on entity pairs in chemical patents. Moreover, we perform a stress test evaluation by simulating OCR failures and NER errors. Furthermore, we improve the robustness against noise of our model by adding simulated OCR failures and NER errors to the training.

%%%%%%%%%%%%%%%%%%%%%%%%%%%%%%%%
\subsection{Relation Classification}
%%%%%%%%%%%%%%%%%%%%%%%%%%%%%%%%

As mentioned in \autoref{subsec:rc}, transformer models achieve state-of-the-art results in several NLP tasks \cite{Zhang2020RetrospectiveComprehension, Bataa2019AnJapanese, Hakala2019BiomedicalBERT, Zhong2019SearchingNext}. 
Likewise, we find that BERT models are the state-of-the-art in the task of relation classification in chemical patents \cite{Dutt2021APatents}.

\begin{figure}[h!]
  \centering
  \includegraphics[width=\linewidth]{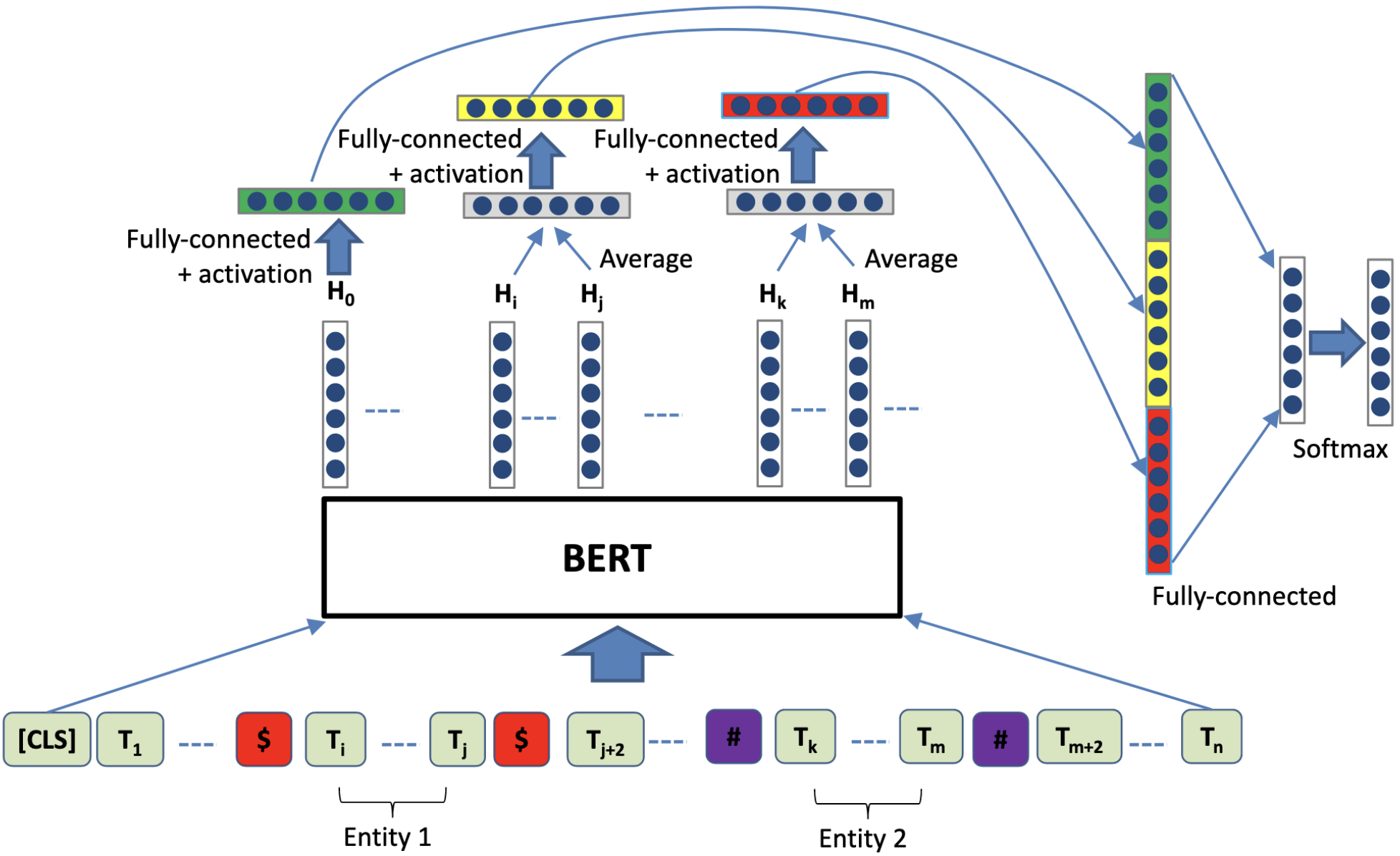}
  \caption{Model architecture \cite{Wu2019EnrichingClassification}.}
  \label{fig:model_arch}
\end{figure}

In this work, we replicate the work of \cite{Dutt2021APatents}. As in \cite{Dutt2021APatents}, we experiment with the following five transformer models: BERT-base \cite{Devlin2019BERT:Understanding}, BERT-large \cite{Devlin2019BERT:Understanding}, BioBERT \cite{Lee2020BioBERT:Mining}, Clinical-BERT \cite{Alsentzer2019PubliclyEmbeddings} and PubMed-BERT \cite{Peng2019TransferDatasets}. 
Transformer models for the task of relation classification have been improved by adding entity markers on the relation classification benchmarks \cite{Soares2019MatchingLearning, Wu2019EnrichingClassification}:  SemEval 2010 Task 8 \cite{Hendrickx2010SemEval-2010Nominals}, KBP-37 \cite{Zhang2015RelationNetwork} and TACRED \cite{Zhang2017Position-awareFilling}. These entity markers give the model knowledge about the location of entities and showed significant improvements in the performance.
Similarly, \cite{Dutt2021APatents} embedded the entities to locate the entities in a sentence. As we are replicating \cite{Dutt2021APatents}, we develop our models using the following entity markers: "\$" to the start and end of the first entity and "\#" to the start and end of the second entity.
Moreover, "[CLS]" is added at the beginning of a sentence. We use the model architecture as in \cite{Wu2019EnrichingClassification} to build our relation classification model (Figure \ref{fig:model_arch}). 

%%%%%%%%%%%%%%%%%%%%%%%%%%%%%%%%
\subsection{Stress Test Evaluation}
%%%%%%%%%%%%%%%%%%%%%%%%%%%%%%%%

For our stress test evaluation, we create several stress test datasets. By doing so, we aim to evaluate whether our model is robust against OCR failures and NER errors.

We mentioned in \autoref{intro} that chemical patents are typically in XML, HTML or image PDF format \cite{Akhondi2014AnnotatedMining}. Chemical patents in the format of XML or HTML are mostly extracted text from image PDFs \cite{Akhondi2014AnnotatedMining}. 
To extract the text from image PDFs, OCR is used \cite{Akhondi2014AnnotatedMining}. However, OCR is likely to introduce errors in their output \cite{Lopresti2009OpticalProcessing}. These errors are mostly typographical errors, due to OCR failures \cite{Sayle2011ImprovedCorrection}. An example of an OCR failure is when the character "i" is extracted as "l". Moreover, the output of OCR could have errors due to human mistakes. An example of a human mistake is when a character "a" is substituted by "s", in which "s" is an adjacent character of "a" on the QWERTY keyboard. 

In this paper, we simplify OCR failures as corrupting random characters. Based on the method used in \cite{Araujo2021StressEmbeddings}, we use two types of errors to simulate typographical errors, namely \textit{keyboard typo noise} and \textit{swap noise}. \textit{Keyboard typo noise} is created by replacing a character with an adjacent character on the QWERTY keyboard. \textit{Swap noise} is created by swapping two adjacent characters.

Furthermore, the input data of our relation classification model can consists of errors made by a NER. 
There are five types of mistakes that a NER model can make \cite{Nejadgholi2020ExtensiveExperience}. First, an entity is predicted by the NER, but is not an annotated entity in the corpus. Second, an an annotated entity is not predicted. Third, an entity is predicted but the label is wrong. Fourth, an entity is predicted with overlapping spans but the label is wrong. Fifth, an entity is predicted with overlapping spans and the label is right. 

In our case, every entity has the same label. Hence, label mismatches do not apply to the NER model for our task. Moreover, we do not focus on the first and second type of mistakes, as the relation classification cannot do anything about it. We focus on the fifth mistake in which an entity is predicted with overlapping spans and has the right label. Such mistakes are also referred as span boundary mistakes, which were a common error of the NER model in the mention extraction phase of anaphora resolution in chemical patents \cite{Dutt2021APatents}.  

%%%%%%%%%%%%%%%%%%%%%%%%%%%%%%%%
\section{Experimental Setup}
%%%%%%%%%%%%%%%%%%%%%%%%%%%%%%%%

%%%%%%%%%%%%%%%%%%%%%%%%%%%%%%%%
\subsection{Data}
%%%%%%%%%%%%%%%%%%%%%%%%%%%%%%%%

We use the ChEMU-Ref dataset \cite{Fang2021ChEMU-Ref:Domain}. The ChEMU-Ref dataset consists of English chemical patents snippets from the European Patent Office and the United States Patent and Trademark Office \cite{Fang2021ChEMU-Ref:Domain}. The entities in the chemical patent snippets are annotated and pairs of entities are linked to an anaphoric relation. The ChEMU-ref dataset has an inner-annotator agreement of Krippendorff's $\alpha = 0.84$ \cite{Fang2021ChEMU-RefDomain}. Figure \ref{fig:AR_example} shows an example of an annotated snippet of a chemical patent. There are five anaphoric relations in chemical patents: co-reference, transformed, reaction associated, work up and contained \cite{Dutt2021APatents, Fang2021ChEMU-Ref:Domain, Li2021ExtendedPatents}. 

The dataset consists of 1125 chemical patent snippets. The patents were are annotated using Brat, which is a web-based text annotation tool \cite{Stenetorp2012BRAT:Annotation}. Annotations created in Brat are stored in a standoff format -- text files (.txt) with the text and corresponding annotation files (.ann) with the annotations \footnote{See Appendix \ref{app:brat} for an example of a text file and corresponding annotation file.}. 

%%%%%%%%%%%%%%%%%%%%%%%%%%%%%%%%
\subsubsection{Data Cleaning and Pre-processing}
%%%%%%%%%%%%%%%%%%%%%%%%%%%%%%%%

We find one empty and one duplicate text file.
Moreover, we find that three annotation files do not have a corresponding text file. We remove these chemical patent snippets from the data, resulting in 1120 chemical patent snippets.

We prepare the data for our model by creating data samples for each entity pair that has an anaphoric relation. This results in 11832 data samples, i.e., there are 11832 entity pairs that have an anaphoric relation. Figure \ref{fig:org_sample} shows an example of a data sample, in which "To the solution of Compound (4) (0.815 g, 1.30 mmol) in THF (4.9 ml)" and "a flask" have a \textit{contained} relation. 

\begin{figure}
  \centering
  \includegraphics[width=0.85\linewidth]{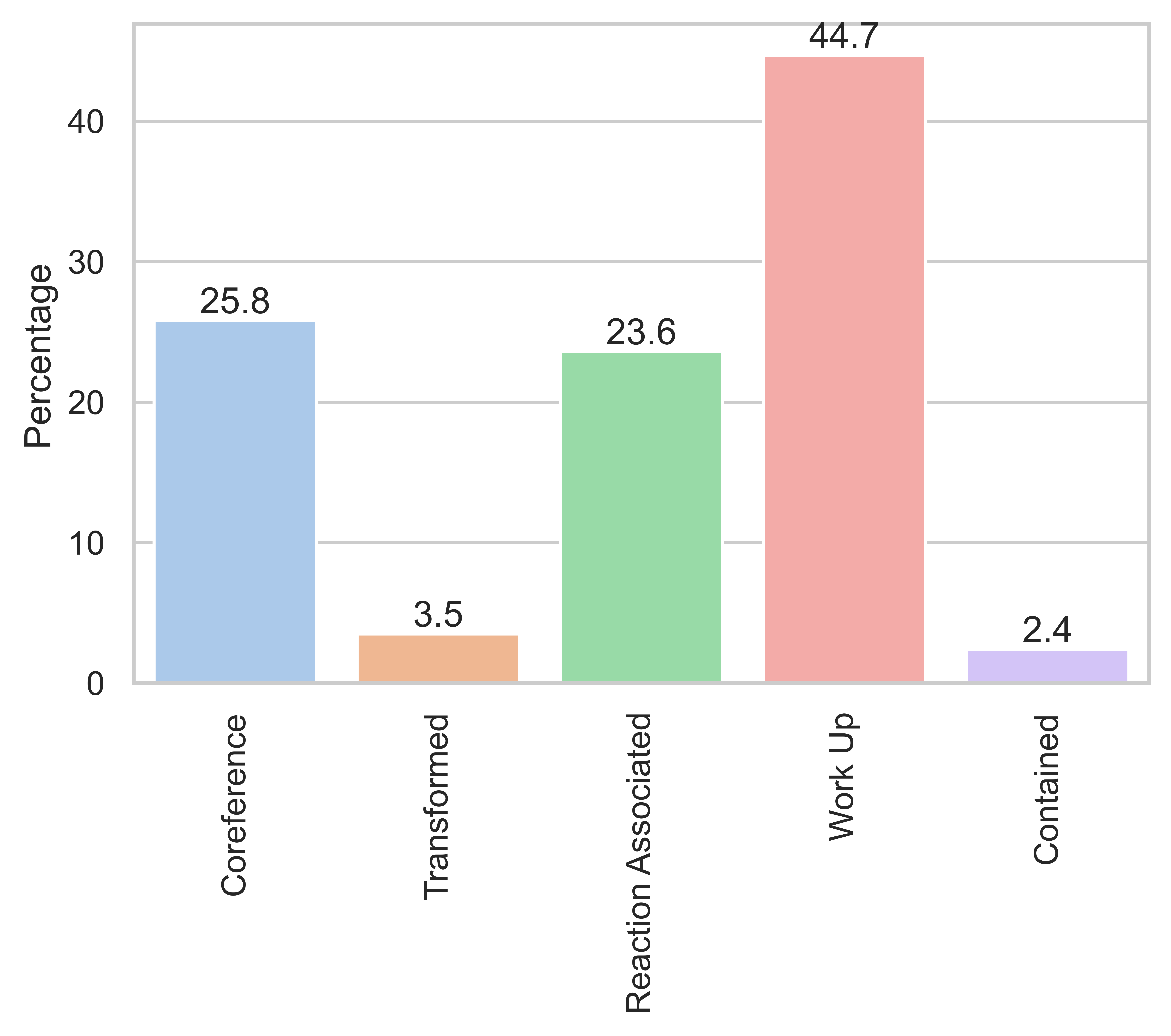}
  \caption{Distribution of the anaphoric relations in all folds and datasets.}
  \label{fig:rel_distr}
\end{figure}

This paper performs a 5-fold cross-validation with a test set. Therefore, we first split the dataset into a train and test set -- 75\% and 25\%, respectively, amounting to a training set size of 8874 data samples and a test size of 2958 data samples.
The train set is used for the 5-fold cross-validation to avoid overfitting. Moreover, we use stratified sampling to ensure that in every fold and dataset the distribution of the relations are the same (Figure \ref{fig:rel_distr}).

We create several input datasets which are pre-processed differently. For our first dataset, which we refer to as our original dataset (\textbf{O}), we add "<e1>" at the start of the first entity and "</e1>" at the end of the first entity. 
Likewise, we add "<e2>" at the start of the second entity and "</e2>" at the end of the second entity. Figure \ref{fig:ent_sample} shows an example of a data sample in our original dataset (\textbf{O}).
The model uses these entity markers to identify the entities in the data samples, and replaces the entity markers of the first entity with "\$" and the second entity with "\#". We do this for the train and test set. 

\begin{figure}
    \centering
\begin{subfigure}{\linewidth}
    \begin{displayquote}

    \textcolor{blue}{To the solution of Compound (4) (0.815 g, 1.30 mmol) in THF (4.9 ml)} in  \textcolor{red}{a flask} were added acetic acid (9.8 ml) and water (4.9 ml).
    
    \end{displayquote}
    \caption{Raw data sample.}
    \label{fig:org_sample}
\end{subfigure}
\label{fig:sample}
\vspace{0.3cm}
\begin{subfigure}{\linewidth}
    \centering
    \begin{displayquote}

    \textbf{<e1>} \textcolor{blue}{To the solution of Compound (4) (0.815 g, 1.30 mmol) in THF (4.9 ml)} \textbf{</e1>} in \textbf{<e2>} \textcolor{red}{a flask} \textbf{</e2>} were added acetic acid (9.8 ml) and water (4.9 ml).
    
    \end{displayquote}
    \caption{Original data sample, with added entity tokens.}
    \label{fig:ent_sample}
\end{subfigure}
        
\caption{(a) shows a raw data sample, (b) shows a data sample of the original datset, with added entity tokens.}
\label{fig:sample}
\end{figure}

Furthermore, we create two types of noisy datasets.
First, we create noisy datasets with simulated OCR failures, referred to as \textbf{OCR}. Our \textbf{OCR} datasets consists of two types of noise, namely \textit{keyboard typo noise} and \textit{swap noise} (see Table \ref{tab:sim_err}). Second, we create noisy datasets with simulated NER errors, referred as \textbf{NER}. Our \textbf{NER} datasets consists of five types of span boundary mistakes, namely \textit{span start left}, \textit{span start right}, \textit{span end left}, \textit{span end right} and \textit{span split} (see Table \ref{tab:sim_err}).

We experiment with different Word Error Rates (WER) for the \textbf{OCR} datasets. Based on the method used in \cite{Bazzo2020AssessingRetrieval} and \cite{Hamdi2022In-DepthLinking}, we choose the following WERs: 5\%, 10\%, 25\% and 50\%. 
A WER of 10\% means that 10\% of the words in the data sample contain an error.
Thus, if there are 20 words in a sentence and the WER is 10\%, then two words will have an error. The \textbf{OCR} datasets are created as follows:
\begin{enumerate}
    \item 
    We randomly choose \textit{n} words of the data sample of the original dataset (\textbf{O}), in which \textit{n} corresponds to the WER (5\%, 10\%, 25\% and 50\%). 
    
    \item For every randomly chosen word in step (1), we randomly choose to apply a \textit{keyboard typo noise} or \textit{swap noise}.
\end{enumerate}

For the \textbf{NER} datasets, we experiment with different percentages of data samples that contain an error -- 25\%, 50\%, 75\% and 100\%.
For example, 50\% means that 50\% of the data samples in the dataset have an error. 
We create the \textbf{NER} datasets as follows: 
\begin{enumerate}
    \item We sample the percentage (25\%, 50\%, 75\% and 100\%) that we want to transform from the original dataset (\textbf{O}), and remove those from the dataset.
    
    \item We randomly choose to transform the first or second entity.
    
    \item We randomly choose between \textit{span start left}, \textit{span start right}, \textit{span end left}, \textit{span end right} and \textit{span split}.
    
    \item We add the noisy data to the dataset.
\end{enumerate}
Note that we only perform one type of noise to a data sample for the \textbf{NER} datasets.

\begin{table*}[h]
  \caption{Types of simulated errors.}
  \label{tab:sim_err}
  \begin{tabular}{lll}
    \toprule
    {\textbf{Error}} & \textbf{Description} & {\textbf{Example}} \\
    \midrule
    No error & - & 
    \makecell[l]{\textbf{<e1>} \textcolor{blue}{The reaction mixture} \textbf{</e1>} was  concentrated under \\ 
    reduced pressure, and then \textbf{<e2>} \textcolor{red}{the resulting residue} \textbf{</e2>} \\
    was purified by flash column chromatography \\
    (MeOH:CH2Cl2=0:100→1:80→1:50).} \\
    \midrule
    
    Keyboard typo noise & 
    \makecell[l]{A random character is replaced by \\ 
    an adjacent character on QWERTY \\ 
    keyboard.} & 
    \makecell[l]{\textbf{<e1>} \textcolor{blue}{The reacti\underline{\textbf{0}}n mixture} \textbf{</e1>} was concentrated under \\
    reduced pressure, and then \textbf{<e2>} \textcolor{red}{the resulting residue} \textbf{</e2>} \\
    was purified by flash column chromatography \\
    (MeOH:CH2Cl2=0:100→1:80→1:50).} \\
    \midrule
    
    Swap noise & 
    \makecell[l]{A random character is swapped \\ with an adjacent character.} & 
    \makecell[l]{\textbf{<e1>} \textcolor{blue}{The r\underline{\textbf{ae}}ction mixture} \textbf{</e1>} was concentrated
    under \\ reduced pressure, and then \textbf{<e2>} \textcolor{red}{the
    resulting residue} \textbf{</e2>}   \\
    was purified by flash column chromatography \\
    (MeOH:CH2Cl2=0:100→1:80→1:50).}\\
    \midrule
    
    Span start left & 
    \makecell[l]{The entity token referring to the \\ start of the entity "<e\#>" is moved \\ one word to the left.} & 
    \makecell[l]{\textbf{<e1>} \textcolor{blue}{The reaction mixture} \textbf{</e1>} was concentrated
    under \\ reduced pressure, and \underline{\textbf{<e2>}} \textcolor{red}{\underline{\textbf{then}} the
    resulting residue} \textbf{</e2>} \\
    was purified by flash column chromatography \\
    (MeOH:CH2Cl2=0:100→1:80→1:50).}\\
    \midrule
    
    Span start right & 
    \makecell[l]{The entity token referring to the \\ start of the entity "<e\#>" is moved \\ one word to the right.} & 
    \makecell[l]{\underline{\textbf{The \textbf{<e1>}}} \textcolor{blue}{ reaction mixture} \textbf{</e1>} was concentrated
    under \\ reduced pressure, and then \textbf{<e2>} \textcolor{red}{the
    resulting residue} \textbf{</e2>}  \\
    was purified by flash column chromatography \\
    (MeOH:CH2Cl2=0:100→1:80→1:50).} \\
    \midrule
    
    Span end right & 
    \makecell[l]{The entity token referring to the \\ end of the entity "</e\#>" is moved \\ one word to the right.} & 
    \makecell[l]{\textbf{<e1>} \textcolor{blue}{The reaction mixture \underline{\textbf{was}}} \underline{\textbf{</e1>}} concentrated
    under \\ reduced pressure,  and then \textbf{<e2>} \textcolor{red}{the
    resulting residue} \textbf{</e2>} \\
    was purified by flash column chromatography \\
    (MeOH:CH2Cl2=0:100→1:80→1:50).}\\
    \midrule
    
    Span end left & 
    \makecell[l]{The entity token referring to the \\ end of the entity "</e\#>" is moved \\ one word to the left.} & 
    \makecell[l]{\textbf{<e1>} \textcolor{blue}{The reaction} \underline{\textbf{</e1>} mixture} was  concentrated
    under \\ reduced pressure, and then \textbf{<e2>} \textcolor{red}{the
    resulting residue} \textbf{</e2>} \\ 
    was purified by flash column chromatography \\
    (MeOH:CH2Cl2=0:100→1:80→1:50).} \\
    \midrule
    
    Span split & \makecell[l]{The entity is randomly split into \\ two and randomly chosen one as \\ the new entity.}& 
    \makecell[l]{\textbf{<e1>} \textcolor{blue}{The reaction mixture} \textbf{</e1>} was  concentrated
    under \\ reduced pressure,  and then \underline{\textbf{the
    resulting <e2>}} \textcolor{red}{residue} \textbf{</e2>} \\
    was purified by flash column chromatography \\
    (MeOH:CH2Cl2=0:100→1:80→1:50).} \\
    \bottomrule
  \end{tabular}
\end{table*}

%%%%%%%%%%%%%%%%%%%%%%%%%%%%%%%%
\subsection{Experiments}
%%%%%%%%%%%%%%%%%%%%%%%%%%%%%%%%

For our experiments, we first compare the five BERT models (BERT-base \cite{Devlin2019BERT:Understanding}, BERT-large \cite{Devlin2019BERT:Understanding}, BioBERT \cite{Lee2020BioBERT:Mining}, Clinical-BERT \cite{Alsentzer2019PubliclyEmbeddings} and PubMed-BERT \cite{Peng2019TransferDatasets}) using our original dataset (\textbf{O}). Our next experiments uses the best BERT-model. Second, we experiment whether OCR failures and NER errors have an influence on the performance. We do this by training on the original training set, and evaluating on the \textbf{OCR} and \textbf{NER} stress test sets. Third, we experiment whether adding simulated OCR failures and NER errors to the training set helps to improve the performance and robustness of our model. 

%%%%%%%%%%%%%%%%%%%%%%%%%%%%%%%%
\subsection{Hyperparameters}
%%%%%%%%%%%%%%%%%%%%%%%%%%%%%%%%

The hyperparameters used in our model are found in Table \ref{tab:hyp}.
Due to limited GPU memory access, we choose a train batch size of 4.

\begin{table}[H]
  \caption{Hyperparameters settings.}
  \label{tab:hyp}
  \begin{tabular}{rl}
    \toprule
    {\textbf{Parameter}} & {\textbf{Setting}} \\
    \midrule
    Train batch size & 4 \\
    Evaluation batch size & 8 \\
    Max sentence length & 384 \\
    Learning rate (adamW) & 2e-05  \\
    Train epochs & 5  \\
    Dropout rate & 0.1 \\
    \bottomrule
  \end{tabular}
\end{table}

%%%%%%%%%%%%%%%%%%%%%%%%%%%%%%%%
\subsection{Evaluation}
%%%%%%%%%%%%%%%%%%%%%%%%%%%%%%%%

We evaluate our models based on three metrics -- precision, recall and macro F1. 
Precision is a measure to determine how many of the predicted positives are actual positives, and is calculated as follows:
\begin{equation}
    \textnormal{Precision} = \frac{\textnormal{True Positive }}{\textnormal{True Positive} + \textnormal{ False Positive}}
\end{equation}
Recall is the proportion of true positives that are correctly predicted as positive, and is defined as: 
\begin{equation}
    \textnormal{Recall} = \frac{\textnormal{True Positive }}{\textnormal{True Positive } + \textnormal{ False Negative}}
\end{equation}
F1 is the harmonic mean of precision and recall. Hence, it seeks a balance between precision and recall. F1 is formulated as:
\begin{equation}
    \textnormal{F}_1 = 2 \times \frac{\textnormal{Precision } \times \textnormal{ Recall}}{\textnormal{Precision } + \textnormal{ Recall}}
\end{equation}
The macro F1 score is the arithmetic mean of the F1 score per class:
\begin{equation}
    \textnormal{Macro F}_1 = \frac{1}{5} \sum_{i=0}^4 \textnormal{F}_1\textnormal{ score}_i
\end{equation}
As the macro F1 score equally accounts all classes, it is a suitable metric for our imbalanced dataset \cite{Grandini2020MetricsOverview}\footnote{See \autoref{app:setup} for a visualisation of our full experimental setup.}.

%%%%%%%%%%%%%%%%%%%%%%%%%%%%%%%%
\section{Results and Analysis}
%%%%%%%%%%%%%%%%%%%%%%%%%%%%%%%%

We investigated how the performance of our relation classification model on the task of anaphora resolution in chemical patents differs in a noise-free and noisy environment. Moreover, we investigated to what extent we can improve the robustness against noisy data of our model. 

%%%%%%%%%%%%%%%%%%%%%%%%%%%%%%%%
\subsection{BERT-based models}
%%%%%%%%%%%%%%%%%%%%%%%%%%%%%%%%

We developed a relation classification model using several BERT-based models (BERT-base \cite{Devlin2019BERT:Understanding}, BERT-large \cite{Devlin2019BERT:Understanding}, BioBERT \cite{Lee2020BioBERT:Mining}, Clinical-BERT \cite{Alsentzer2019PubliclyEmbeddings} and PubMed-BERT \cite{Peng2019TransferDatasets}) to classify the different anaphoric relations in chemical patents. The 5-fold cross-validation results of these BERT models are shown in \autoref{tab:bert_cv}. We show that all five BERT models have an average macro F1 score of 0.95. Moreover, the standard deviation of our models shows that there is a low variance ($<0.01$) in the performance of the models.

\begin{table}
  \caption{Results in macro F1 of the BERT models of the 5-fold cross-validation using original train dataset (O). We report the mean and standard deviation of the five validation sets.}
  \label{tab:bert_cv}
  \begin{tabular}{rcccc}
    \toprule
    {\textbf{Model}} & \textbf{Mean} & {\textbf{Min}} & \textbf{Max} & \textbf{Std} \\
    \midrule
    BERT-base & 0.948 & 0.944 & 0.956 & 0.005\\
    BERT-large & 0.950 & 0.946 & 0.953 & 0.003 \\
    BioBERT &  0.953 & 0.944 & 0.970 & 0.010 \\
    Clinical-BERT & 0.950 & 0.947 & 0.956 & 0.004 \\
    PubMed-BERT & 0.946 & 0.935 & 0.954 & 0.007 \\
    \bottomrule
  \end{tabular}
\end{table}

\begin{figure}
    \centering
\begin{subfigure}{\linewidth}
    \includegraphics[width=0.9\textwidth]{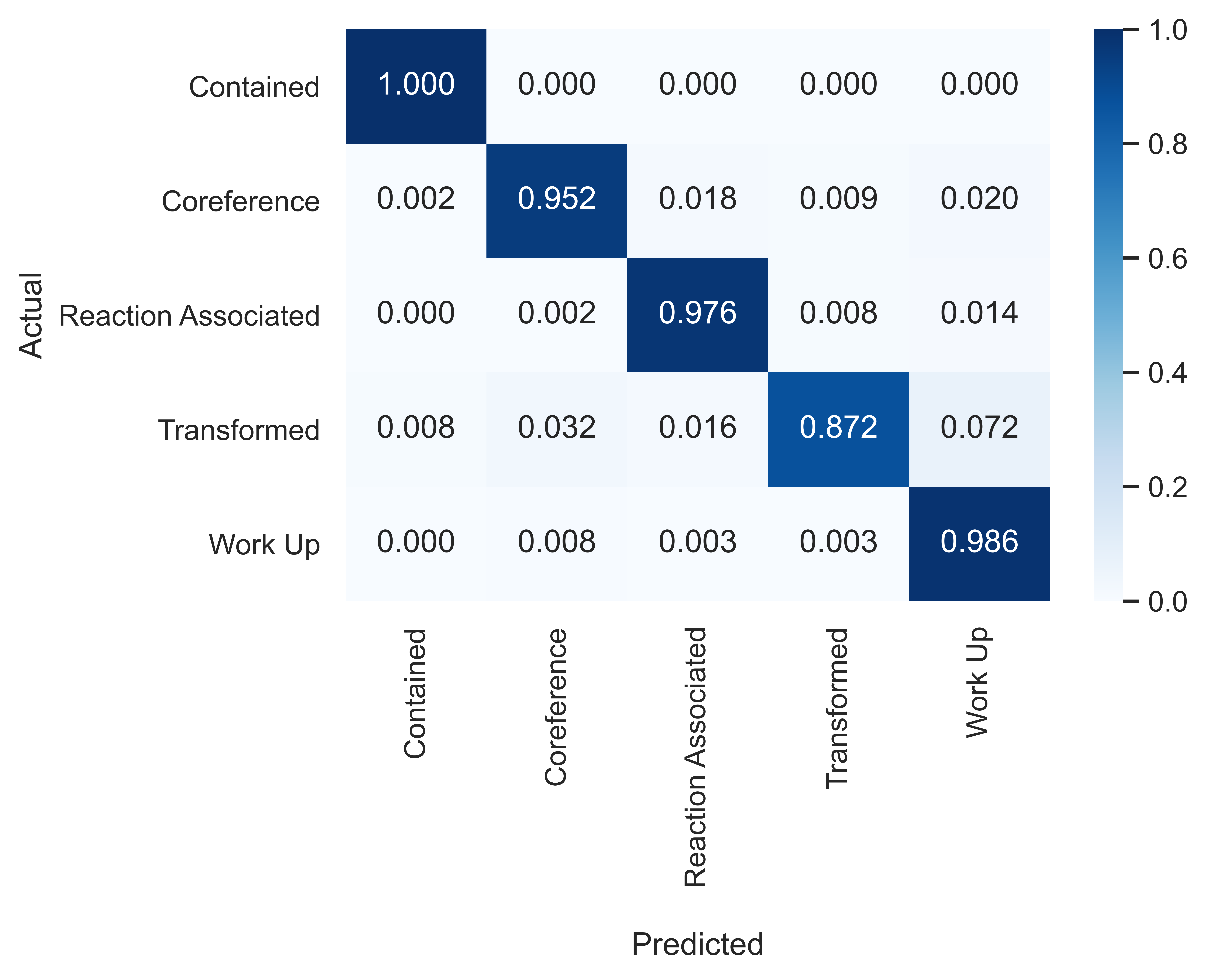}
  \caption{Original test set.}
  \label{fig:cm_bert_test}
\end{subfigure}
\label{fig:cm}

\begin{subfigure}{\linewidth}
    \centering
    \includegraphics[width=0.9\linewidth]{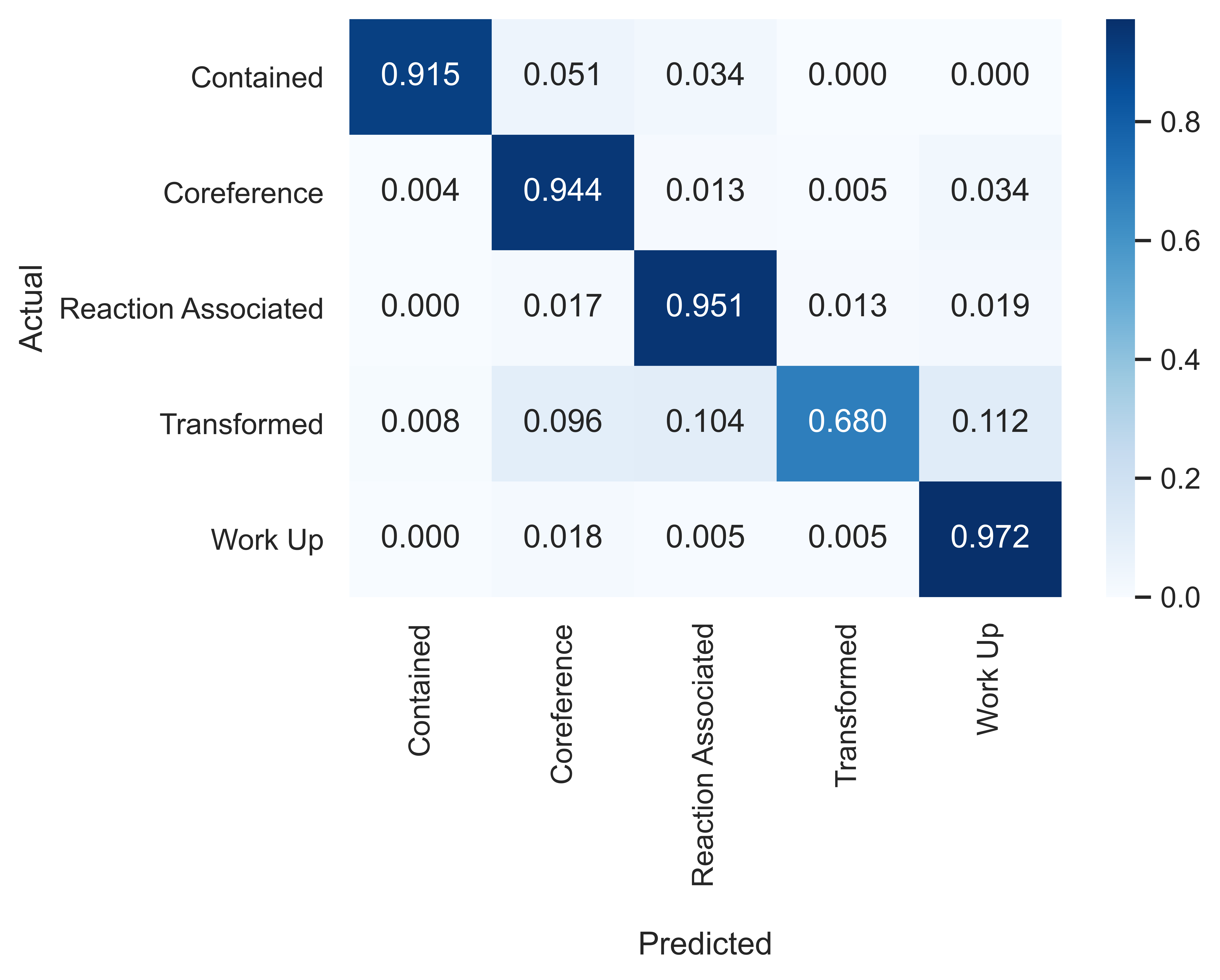}
  \caption{OCR stress test set -- 50\% Word Error Rate.}
  \label{fig:cm_bert_test_ocr}
\end{subfigure}
\label{fig:cm}

\begin{subfigure}{\linewidth}
    \centering
    \includegraphics[width=0.9\linewidth]{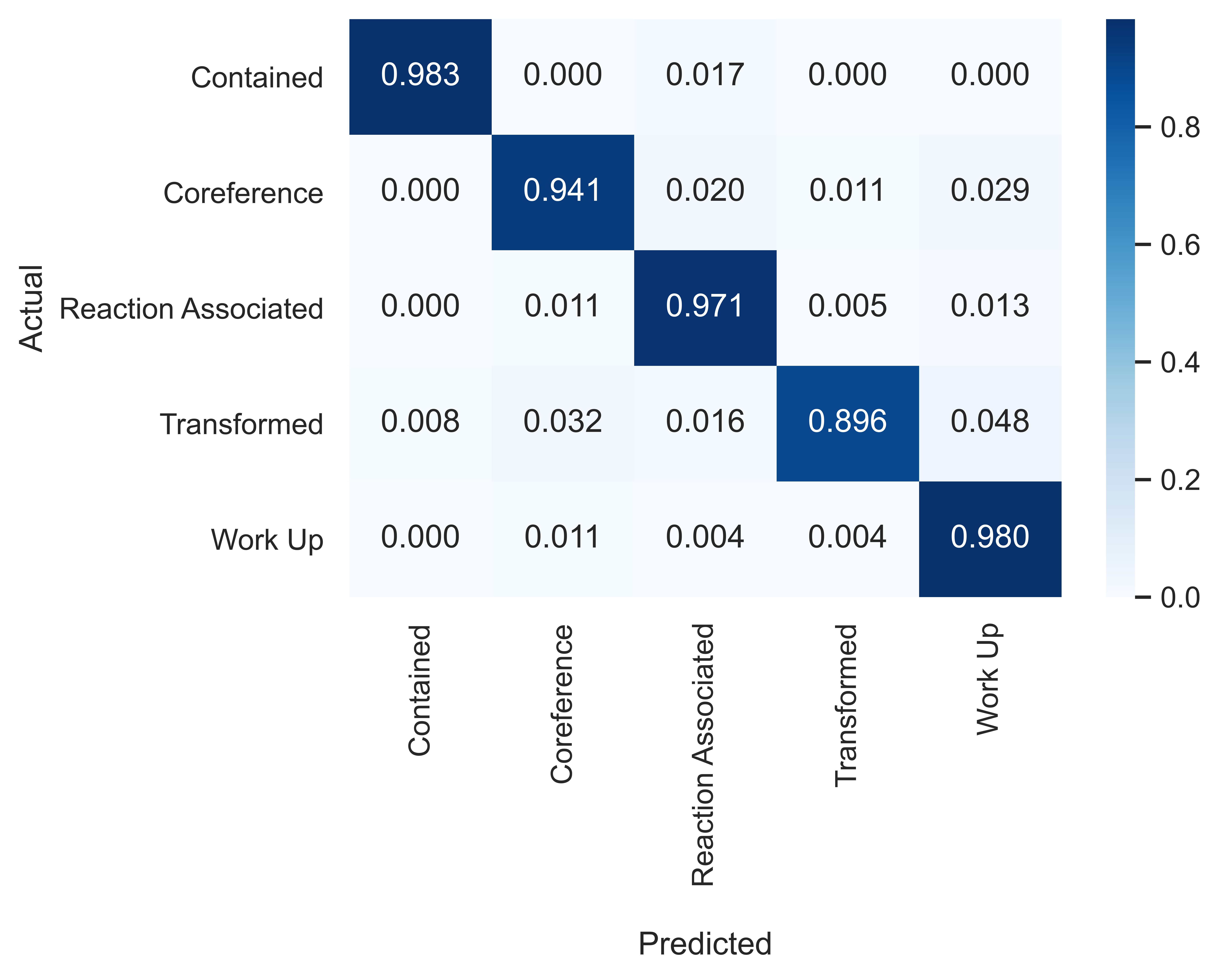}
    \caption{NER stress test set -- 100\%}
    \label{fig:cm_bert_test_ner}
\end{subfigure}
\caption{Normalised confusion matrices of \textit{BERT-base} on different test sets.}
\label{fig:cm}
\end{figure}

As the models have similar performance scores, we choose to experiment further with \textit{BERT-base}, as this model has the lowest amount of parameters. Our \textit{BERT-base} model achieves a macro F1 score of 0.956 on the test set (Table \ref{tab:bert_test}). 
The confusion matrix of \textit{BERT-base} can be seen in \autoref{fig:cm_bert_test}, which shows that most misclassifications come from the \textit{transformed} class.
This might be due to the imbalance of our classes in the dataset -- \textit{transformed} being one of our smallest classes (Figure \ref{fig:rel_distr}).

\begin{figure}
    \centering
\begin{subfigure}{\linewidth}
    \includegraphics[width=\linewidth]{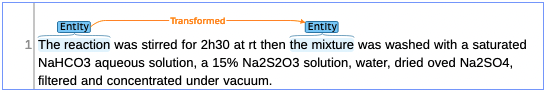}
  \caption{True relation}
  \label{fig:error2_true}
\end{subfigure}
\vspace{0.3cm}
\begin{subfigure}{\linewidth}
    \centering
    \includegraphics[width=\linewidth]{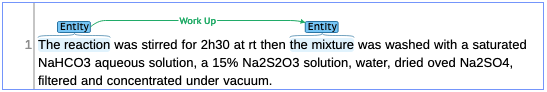}
  \caption{Predicted relation.}
  \label{fig:error2_pred}
\end{subfigure}
\caption{Example of misclassification of \textit{BERT-base}.}
\label{fig:error2}
\end{figure}

An interesting finding is that, even though \textit{contained} is the smallest class in our dataset, our model correctly predicts all \textit{contained} relations. A possible reason is that the signal of \textit{transformed} relationships are weaker than the signal of \textit{contained} relationships. Figure \ref{fig:error2} shows an example, in which \textit{work up} was predicted, but its true relation is \textit{transformed}. In this example, "the mixture" has undergone a transformation, as "the reaction" is \textit{stirred}. However, the model might have been confused as a result of the rest of the sentence, which indicates a \textit{work up} relation. 

\begin{figure}
    \centering
\begin{subfigure}{\linewidth}
    \includegraphics[width=\linewidth]{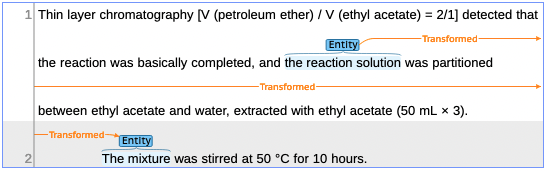}
  \caption{True relation.}
  \label{fig:error1_true}
\end{subfigure}
\vspace{0.3cm}
\begin{subfigure}{\linewidth}
    \centering
    \includegraphics[width=\linewidth]{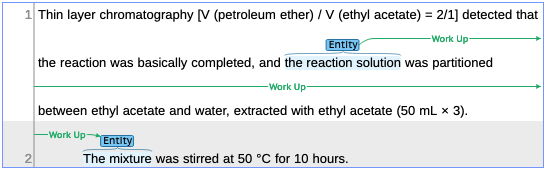}
  \caption{Predicted relation.}
  \label{fig:error1_pred}
\end{subfigure}
\caption{Example of misclassification of \textit{BERT-base}.}
\label{fig:error1}
\end{figure}

Another example is shown in Figure \ref{fig:error1}, in which the true value is \textit{transformed} and the predicted value is \textit{work up}. This example shows that the annotation is not perfect, as the two entities have more a \textit{work up} relation rather than a  \textit{transformed} relation. One should note that the ChEMU-ref dataset had an inner-annotator agreement of Krippendorff's $\alpha = 0.84$ \cite{Fang2021ChEMU-RefDomain}, and therefore annotations of our corpus are not perfect.

\begin{table}
  \caption{Results of \textit{BERT-Base} on test set using original dataset (O).}
  \label{tab:bert_test}
  \begin{tabular}{r|ccc}
    \toprule
    {\textbf{Relation}} & \textbf{Precision} & {\textbf{Recall}} & \textbf{F1} \\
    \midrule
    Contained & 0.967 & 1.000 & 0.983 \\
    Coreference & 0.969 & 0.952 & 0.960  \\
    Reaction Associated &  0.973 & 0.976 & 0.975  \\
    Transformed & 0.879 & 0.872 & 0.876  \\
    Work Up & 0.982 & 0.986 & 0.984  \\
    \midrule
    Macro-average   &   0.954   &  0.957   &  0.956 \\
    \bottomrule
  \end{tabular}
\end{table}

%%%%%%%%%%%%%%%%%%%%%%%%%%%%%%%%
\subsection{Stress Test Evaluation}
%%%%%%%%%%%%%%%%%%%%%%%%%%%%%%%%

We perform a stress test evaluation on our model using \textit{BERT-base}. 

Table \ref{tab:stres_test2} shows the results of the \textbf{OCR} stress test datasets on different word error rates. For all word error rates, the performance of our model drops in comparison to the test set with no errors. We observe that a 5\% and 10\% WER decreases the macro F1 score by 1.3\%, 25\% WER by 5.6\% and 50\% WER  by 6.9\%. 
The confusion matrix of the \textbf{OCR} stress test dataset with a WER of 50\% can be seen in Figure \ref{fig:cm_bert_test_ocr}, in which we show that the drop in performance mainly comes from the misclassifications of the \textit{transformed} relationship. 
Our model evaluated on our original test dataset (\textbf{O}) showed that the majority of the misclassifications were caused by the \textit{transformed} relation. 
Hence, it may not come as a surprise that the drop in performance of the \textbf{OCR} stress test dataset mainly comes from the misclassifications of the \textit{transformed} relation. 

The results of the \textbf{NER} stress test datasets are shown in Table \ref{tab:stres_test}. One should note that the percentages refer to the amount of data samples consisting a NER error. 
We observe that the 25\% NER stress test dataset decreases the macro F1 score by 0.3\%, 75\% by 0.8\% and 100\% by 0.4\%. Interestingly, we observe that the performance increases by 0.1\% on the 50\% stress test dataset. In general, the performance drops on the \textbf{NER} stress test datasets, in comparison to the original test dataset (\textbf{O}), are rather small. These results indicate that simulated \textbf{NER} errors do not have a strong influence on the performance of the model. 
Figure \ref{fig:cm_bert_test_ner} shows the confusion matrix of the 100\% NER stress test dataset, which shows similar results as our original test dataset (\autoref{fig:cm_bert_test}). 

\begin{table}
  \caption{Results in macro F1 of OCR stress test evaluation experiments. We report the mean and standard deviation by training and evaluating five times with different seeds.}
  \label{tab:stres_test2}
  \begin{tabular}{rcccc}
    \toprule
    & \multicolumn{4}{c}{\textbf{Word Error Rate}}\\
    \cmidrule(lr){2-5}
    {\textbf{Error Type}} & \textbf{5\%}& {\textbf{10\%}} & \textbf{25\%} & \textbf{50\%}\\
    \midrule
    OCR & \makecell[l]{0.944 \\ {\footnotesize $\pm$0.003}}  & \makecell[l]{0.944 \\ {\footnotesize $\pm$0.006}} & \makecell[l]{0.902 \\ {\footnotesize $\pm$0.011}} & \makecell[l]{0.890 \\ {\footnotesize $\pm$0.049}} \\
    \bottomrule
  \end{tabular}
\end{table}

\begin{table}
  \caption{Results in macro F1 of NER stress test evaluation experiments. We report the mean and standard deviation by training and evaluating five times with different seeds. Percentage reflects on the amount of data samples that have an error.}
  \label{tab:stres_test}
  \begin{tabular}{rcccc}
    \toprule
    & \multicolumn{4}{c}{\textbf{Percentage}}\\
    \cmidrule(lr){2-5}
    {\textbf{Error Type}} & \textbf{25\%}& {\textbf{50\%}} & \textbf{75\%} & \textbf{100\%}\\
    \midrule
    NER & \makecell[l]{0.953 \\ {\footnotesize $\pm$0.005}}  & \makecell[l]{0.957 \\ {\footnotesize $\pm$0.004}} & \makecell[l]{0.948 \\ {\footnotesize $\pm$0.001}} & \makecell[l]{0.952 \\ {\footnotesize $\pm$0.003}} \\
    \bottomrule
  \end{tabular}
\end{table}

%%%%%%%%%%%%%%%%%%%%%%%%%%%%%%%%
\subsection{Training on Simulated Errors}
%%%%%%%%%%%%%%%%%%%%%%%%%%%%%%%%

We introduce simulated OCR and NER errors to the training to evaluate to what extent we can improve the robustness of our model. Note that the increases and decreases of the performance are in macro F1 and in comparison to our original training dataset (\textbf{O}).

We conducted experiments with different word error rates of the OCR errors added to the training (see Table \ref{tab:exp3a}).  
In general, we observe that training with simulated OCR errors decreases the performance on the original test dataset (\textbf{O}). Our results show that adding 5\% WER to the training decreases the performance on the original test dataset (\textbf{O}) by 0.6\%, 10\% by 0.2\%, 25\% by 0.5\% and 50\% by 4.2\%. 
Furthermore, we observe that for the OCR stress test dataset with 5\% WER, adding 5\% WER of OCR errors to the training increases the performance the most, with an increase of 0.5\%.  
Moreover, evaluating the OCR stress test with 10\% WER, we show that the largest increase of performance (0.4\%) is found by adding 10\% WER of OCR to the training. 
For the OCR stress test with 25\% WER, we find that adding 25\% WER of OCR errors to the training increases the performance the most by 3.8\%. 
The performance of the OCR stress test with 50\% is improved the most by adding 25\% WER of OCR errors to the training, resulting in an increase of 3.6\%.
We find that adding 50\% WER of OCR errors to the training does not make the relation classification model more robust. We believe that this is due to adding too much errors to the training, which might have caused the model learning wrong signals.
In general, we find that adding OCR errors to the training lowers the drop in performance on the \textbf{OCR} stress test datasets. 

Adding simulated NER errors to the training shows mixed results (Table \ref{tab:exp3b}). 
We observe that adding an NER error to 25\% of the training data samples improves the models performance on the original test dataset (\textbf{O}), 25\% and 100\% \textbf{NER} stress test dataset by 0.4\%, 0.6\% and 0.2\%, respectively.
For the 50\% \textbf{NER} stress test dataset, we find that adding \textbf{NER} errors to the dataset does not improve the model's performance. The largest improvement in performance for the 75\% \textbf{NER} stress test dataset is found by adding an NER error to 75\% of the training data samples (0.6\%).

\begin{table}
  \caption{Results in macro F1 of training with added simulated OCR  errors on different stress test sets. Percentage refers to the WER. O stands for our original datasets.}
  \label{tab:exp3a}
  \begin{tabular}{rcccccc}
    \toprule
    &  \multicolumn{5}{c}{\textbf{Test}}\\
    \cmidrule(lr){2-6} 
    {\textbf{Train}} &  \textbf{O} & \textbf{5\%}& {\textbf{10\%}} & \textbf{25\%} & \textbf{50\%}\\
    \cmidrule(lr){1-6}
    
     \textbf{O} & 0.956 & 0.944 & 0.944 & 0.902 & 0.890   \\
   
     \textbf{5\%} & 0.950 & \textbf{0.949} & 0.947 & 0.919 & 0.890 \\
    
     \textbf{10\%} & 0.954 & 0.944 & \textbf{0.948} & 0.930 &  0.907 \\
    
     \textbf{25\%} & 0.951 & 0.939 & 0.936 & \textbf{0.937} & \textbf{0.922} \\
    
     \textbf{50\%} & 0.916 & 0.914 & 0.913 & 0.920 & 0.912 \\
    \bottomrule
  \end{tabular}
\end{table}

\begin{table}
  \caption{Results in macro F1 of training with added simulated NER errors on different stress test sets. Percentage refers to the amount of data samples with a NER error. O stands for our original datasets.}
  \label{tab:exp3b}
  \begin{tabular}{rcccccc}
    \toprule

    & \multicolumn{5}{c}{\textbf{Test}}\\
    \cmidrule(lr){2-6}
    {\textbf{Train}} &  \textbf{O} & \textbf{25\%}& {\textbf{50\%}} & \textbf{75\%} & \textbf{100\%}\\
    \midrule
    
     \textbf{O} & 0.956 & 0.953 & 0.957 & 0.948 & 0.952   \\
   
     \textbf{25\%} & 0.960 & \textbf{0.959} & 0.954 & 0.952 & \textbf{0.954} \\
    
     \textbf{50\%} & 0.950 & 0.949 & 0.955 & 0.946 &  0.945 \\
    
     \textbf{75\%} & 0.952 & 0.952 & 0.953 & \textbf{0.954} & 0.948 \\
    
     \textbf{100\%} & 0.945 & 0.943 & 0.946 & 0.947 & 0.943 \\
    \bottomrule
  \end{tabular}
\end{table}

%%%%%%%%%%%%%%%%%%%%%%%%%%%%%%%%
\section{Discussion and Limitations}
\label{disc}
%%%%%%%%%%%%%%%%%%%%%%%%%%%%%%%%

The goal of this paper was to investigate to what extent our relation classification model's performance differs in a noise-free and noisy environment and to what extent we can improve the robustness of our model against noisy data. 

To investigate this, we developed a relation classification model using \textit{BERT-base}, performed a stress test evaluation to evaluate how our model performs when simulated OCR failures and NER errors are introduced and improve the model's robustness by introducing simulated OCR failures and NER errors to the training.

Our relation classification model achieved a macro F1 score of 0.96 on the ChEMU-ref dataset in a noise-free environment.
A limitation of our relation classification model is that we did not include negative samples, i.e., a \textit{no relation} class. As a result, our study replicates a perfect setting, in which we know all entity pairs and their corresponding relation. In a real-world setting, however, we cannot assume that every entity pair, found by the named entity recognition model, has an anaphoric relation. Including a \textit{no relation} class might have an influence on the performance of the model, and is possibly the reason that our model performs better than  \cite{Dutt2021APatents}. 

In general, our stress test evaluation shows that  simulated OCR failures and NER errors drop the model's performance. 
We found a low variance evaluating our stress tests on five different seeds.
Moreover, the largest drop in performance was observed on the \textbf{OCR} stress test evaluations. Hence, our results showed that if no proper corrections are made that OCR failures could have a negative impact on the performance of the relation classification task. 

The NER errors showed a small drop in performance, indicating that span boundary mistakes do not have a major impact on the relation classification. However, there are other NER errors that we did not capture such as wrongly predicted entities and not predicted entities by the NER \cite{Nejadgholi2020ExtensiveExperience}.
We did not capture such errors since the relation classification model cannot do anything about it. However, such errors will likely have a negative impact on the model, emphasising the importance of optimising the named entity recognition model in the mention extraction phase.

A limitation of our stress test evaluation is that we only evaluated the robustness of \textit{BERT-base}. However, as \cite{Dutt2021APatents} found that BERT-based models trained on clinical data, i.e., PubMed-BERT and BioBERT, performed slightly better, it might be interesting to evaluate how robust these models are. 

Another limitation of our stress test evaluation is how we simulated OCR failures. We simplified OCR failures as typographical errors, i.e., adding errors such as \textit{keyboard typo noise} and \textit{swap noise}. Hence, we randomly corrupted characters. However, more specific OCR failures are \textit{substitution errors} ("i" $\rightarrow$ "l"), \textit{space deletion} ("the mixture $\rightarrow$ "themixture"), \textit{space insertion} ("flask" $\rightarrow$ "fl ask"), \textit{character deletion} ("reaction" $\rightarrow$ "reacton") and \textit{character insertion} ("residue" $\rightarrow$ "ressidue") \cite{Lopresti2009OpticalProcessing, Hamdi2022In-DepthLinking}.

Moreover, our results seem to indicate that adding simulated OCR errors with a WER up to 25\% to the training may improve the robustness of the relation classification task. 
As the NER errors did not drop the model's performance on the \textbf{NER} stress test datasets, we did not expect a large decrease in the performance drop by adding simulated NER errors. Interestingly, we found that adding a NER error to 25\% of the data samples in the training and test set increased the performance slightly. 
However, we did not evaluate these experiments on different seeds due to time limitations. Hence, no conclusion can be made whether a significant increase or decrease in performance is found. 

Our work indicates that the performance of the relation classification task in anaphora resolution is negatively impacted by OCR failures. However, we show that the robustness can be improved by adding simulated OCR failures to the training. Moreover, our findings suggest that NER errors, i.e., span boundary mistakes, only have a small impact on the model, indicating that our model is robust against span boundary mistakes.

%%%%%%%%%%%%%%%%%%%%%%%%%%%%%%%%
\section{Conclusions and Future work}
%%%%%%%%%%%%%%%%%%%%%%%%%%%%%%%%

Automating information extraction from chemical patents is crucial due to the vast number of available chemical patents \cite{Li2021ExtendedPatents, Akhondi2019AutomaticPatents} and the importance of timely acquisition of their information \cite{Dutt2021APatents, Zhai2019ImprovingEmbeddings}.  An important component of comprehensive information extraction is anaphora resolution \cite{Rosiger2019ComputationalResolution, Fang2021ChEMU-Ref:Domain}. 

This research aims to answer to what extent the relation classification model on the task of anaphora resolution in chemical patents differs in a noise-free and noisy environment and to what extent we can improve the robustness of our model against OCR failures and NER errors.

To answer this, we develop a relation classification model using \textit{BERT-base}, achieving a macro F1 score of 0.96 in a noise-free environment. 
Moreover, we perform a stress test evaluation to evaluate whether our model is robust against OCR failures and NER errors. Hence, simulating a noisy environment. Our stress test evaluation indicated that OCR failures negatively impacted the model's performance, suggesting that the relation classification on the task of anaphora resolution is not robust against OCR failures. 
Future work could examine how the model responds when OCR failures are simulated in a more advanced way. In other words, simulating OCR failures based on common mistakes by the OCR, such as substitution errors, space deletion and insertion, and character deletion and insertion \cite{Lopresti2009OpticalProcessing, Hamdi2022In-DepthLinking}.

Moreover, the stress test evaluation on the NER errors shows a small performance drop, indicating that span boundary mistakes do not have a major influence on the performance of the relation classification.
We mentioned that there are other NER errors, such as falsely predicted entities and entities that should have been predicted but are not \cite{Nejadgholi2020ExtensiveExperience}. We do not capture these errors as the relation classification cannot do anything about such errors. However, this emphasises the importance of an optimised named entity recognition model, as such errors will impact the performance of the anaphora resolution task.

Furthermore, we improve the robustness against OCR failures on the relation classification task by adding simulated OCR failures to the training. Therefore, we show that adding simulated OCR failures improves the robustness against OCR failures. As the NER errors do not have a major influence on the performance, we find that adding NER errors to the training does not improve the robustness of the model. 
However, the experiments on training with simulated OCR failures and NER errors are not tested on different seeds due to time limitations. Therefore, future work could investigate this to exclude randomness and improve generalisability. 

%%%%%%%%%%%%%%%%%%%%%%%%%%%%%%%%
\section{Ethical Considerations and Aknowledgements}
%%%%%%%%%%%%%%%%%%%%%%%%%%%%%%%%

This work uses the ChEMU-Ref dataset  \cite{Fang2021ChEMU-Ref:Domain}. The ChEMU-Ref dataset was created for the task of anaphora resolution in the ChEMU Shared Task 2022. The dataset is publicly available, however, may only be used for the shared task, educational or research purposes. 
This work was partially supported by the ChEMU project
, funded jointly by the Australian Research Council (Linkage Project LP160101469) and Elsevier.

\bibliographystyle{ACM-Reference-Format}
\bibliography{references}
\onecolumn
\newpage
\appendix

%%%%%%%%%%%%%%%%%%%%%%%%%%%%%%%%
\begin{appendices}
%%%%%%%%%%%%%%%%%%%%%%%%%%%%%%%%

%%%%%%%%%%%%%%%%%%%%%%%%%%%%%%%%
\section{Visualisation Experimental Setup}
\label{app:setup}
%%%%%%%%%%%%%%%%%%%%%%%%%%%%%%%%

\vspace{\fill}

\begin{figure*}[h]
  \centering
  \includegraphics[width=\linewidth]{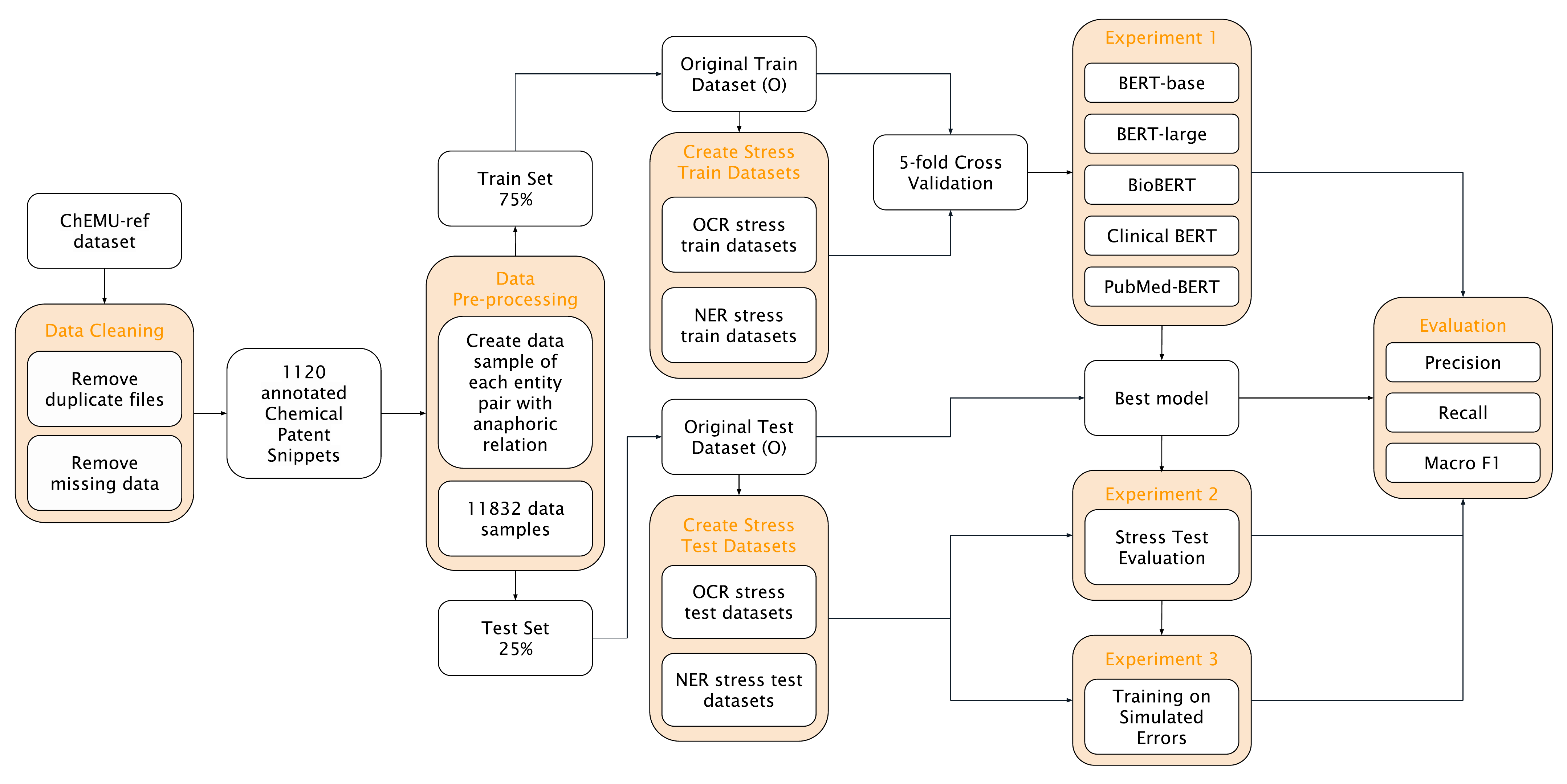}
  \caption{Experimental setup of this work.}
  \label{fig:setup}
\end{figure*}

\vspace{\fill}

\newpage

%%%%%%%%%%%%%%%%%%%%%%%%%%%%%%%%
\section{Brat Standoff Format}
\label{app:brat}
%%%%%%%%%%%%%%%%%%%%%%%%%%%%%%%%

\vspace{\fill}

\begin{figure}[h!]
    \centering
    \begin{tcolorbox}
    PREPARATIVE EXAMPLE 2
    
    Preparation of (S)-5-chloro-3-phenyl-2-(pyrrolidine-2-yl)quinazoline-4(3H)-one
    
    Step 1: Preparation of tert-butyl (S)-2-(5-chloro-4-oxo-3-phenyl-3,4-dihydroquinazoline-2-yl)pyrrolidine-1-carboxylate
    
    5.51 g of tert-butyl (S)-2-(5-chloro-4-oxo-3-phenyl-3,4-dihydroquinazoline-2-yl)pyrrolidine-1-carboxylate was prepared as a beige solid by using 3.76 g (17.47 mmol) of (tert-butoxycarbonyl)-L-proline according to the same manner as described in step 2 of Preparative Example 1 (12.94 mmol, yield: 74\%).
        
    \end{tcolorbox}
\caption{Example of a TXT-file of a chemical patent snippet.}
\label{fig:txt}
\end{figure}

\vspace{\fill}

\begin{table*}[h]
  \caption{Example of corresponding ANN-file of chemical patent snippet as in figure \ref{fig:txt}. In the first column: T\# stands for a text-bound annotation, and R\# stands for a relation.}
  \label{tab:ann}
  \begin{tabular}{cll}
    \toprule
     & \textbf{Location/Relation} & \textbf{Text} \\
    \midrule
    T1 & ENTITY 342 355 & a beige solid \\
    T2 & ENTITY 124 219 & \makecell[l]{tert-butyl (S)-2-(5-chloro-4-oxo-3-phenyl-3,\\4-dihydroquinazoline-2-yl)pyrrolidine-1-carboxylate} \\
    T3 & ENTITY 220 325 &  \makecell[l]{5.51 g of tert-butyl (S)-2-(5-chloro-4-oxo-3-phenyl-3,\\4-dihydroquinazoline-2-yl)pyrrolidine-1-carboxylate} \\
    T4 & COREFERENCE 342 355 & a beige solid\\
    T5 & ENTITY 365 419 & 3.76 g (17.47 mmol) of (tert-butoxycarbonyl)-L-proline \\
    T6 & REACTION\_ASSOCIATED 498 520 & 12.94 mmol, yield: 74\%\\
    T7 & COREFERENCE 220 325 &  \makecell[l]{5.51 g of tert-butyl (S)-2-(5-chloro-4-oxo-3-phenyl-3,\\4-dihydroquinazoline-2-yl)pyrrolidine-1-carboxylate} \\
    T8 & COREFERENCE 498 520 & 12.94 mmol, yield: 74\% \\
    R1 & REACTION\_ASSOCIATED Arg1:T6 Arg2:T5 & \\
    R2 & COREFERENCE Arg1:T7 Arg2:T2& \\
    R3 & COREFERENCE Arg1:T4 Arg2:T3 & \\
    R4 & COREFERENCE Arg1:T8 Arg2:T1 & \\
    R5 & COREFERENCE Arg1:T4 Arg2:T2 & \\
    R6 & COREFERENCE Arg1:T8 Arg2:T2 & \\
    R7 & COREFERENCE Arg1:T8 Arg2:T3 & \\
  \bottomrule
\end{tabular}
\end{table*}

\vspace{\fill}

%%%%%%%%%%%%%%%%%%%%%%%%%%%%%%%%
\end{appendices}
%%%%%%%%%%%%%%%%%%%%%%%%%%%%%%%%

%%%%%%%%%%%%%%%%%%%%%%%%%%%%%%%%
\end{document}